\documentclass[11pt,a4paper]{article}
\usepackage[hyperref]{acl2021}
\usepackage{times}
\usepackage{latexsym}

\usepackage{microtype}

\aclfinalcopy 

\usepackage{multirow}
\usepackage{booktabs}
\usepackage{graphicx}
\usepackage{amsmath}
\usepackage{graphicx}
\usepackage{epstopdf}
\usepackage{subfigure}
\usepackage{amssymb}
\usepackage{color}

\title{Code Summarization with Structure-induced Transformer}

\author{
	Hongqiu Wu\textsuperscript{\rm 1,2,3},
	Hai Zhao\textsuperscript{\rm 1,2,3, \thanks{Corresponding author. This paper was partially supported by National Key Research and Development Program of China (No. 2017YFB0304100), Key Projects of National Natural Science Foundation of China (U1836222 and 61733011), Huawei-SJTU long term AI project, Cutting-edge Machine Reading Comprehension and Language Model. This work was supported by Huawei Noah's Ark Lab.}},
	Min Zhang\textsuperscript{\rm 4} \\
	\textsuperscript{\rm 1} Department of Computer Science and Engineering, Shanghai Jiao Tong University \\
	\textsuperscript{\rm 2} Key Laboratory of Shanghai Education Commission for Intelligent Interaction \\
	and Cognitive Engineering, Shanghai Jiao Tong University, Shanghai, China\\
	\textsuperscript{\rm 3} MoE Key Lab of Artificial Intelligence, AI Institute, Shanghai Jiao Tong University, Shanghai, China \\
	\textsuperscript{\rm 4} Institute of Artificial Intelligence, School of Computer Science and Technology, 
	Soochow University, Suzhou, China \\
	{\texttt wuhongqiu@sjtu.edu.cn, zhaohai@cs.sjtu.edu.cn, minzhang@suda.edu.cn} \\
}

\begin{document}

\maketitle

\begin{abstract}
Code summarization (CS) is becoming a promising area in recent language understanding, which aims to generate sensible human language automatically for programming language in the format of source code, serving in the most convenience of programmer developing. It is well known that programming languages are highly structured. Thus previous works attempt to apply structure-based traversal (SBT) or non-sequential models like Tree-LSTM and graph neural network (GNN) to learn structural program semantics. However, it is surprising that incorporating SBT into advanced encoder like Transformer instead of LSTM has been shown no performance gain, which lets GNN become the only rest means modeling such necessary structural clue in source code. To release such inconvenience, we propose structure-induced Transformer, which encodes sequential code inputs with multi-view structural clues in terms of a newly-proposed structure-induced self-attention mechanism. Extensive experiments show that our proposed structure-induced Transformer helps achieve new state-of-the-art results on benchmarks.
\end{abstract}

\section{Introduction}

\begin{table}[t]
\centering
\resizebox{0.48\textwidth}{!}{%
\begin{tabular}{l|l}
\hline\hline
\begin{tabular}[c]{@{}l@{}}Code\\ (Java)\end{tabular} & \begin{tabular}[c]{@{}l@{}}\textcolor[rgb]{0.5,0.25,0.6}{private} \textcolor[rgb]{0.15,0.5,0.4}{void} \textcolor{blue}{attachPlot} (SVGPlot newplot) \{\\
\quad\; \textcolor[rgb]{0.5,0.25,0.6}{this}.plot = newplot;\\
\quad\; \textcolor[rgb]{0.5,0.25,0.6}{if} (newplot == null) \{\\
\quad\; \quad\; \textcolor[rgb]{0.5,0.25,0.6}{super}.setSVGDocument(null);\\
\quad\; \quad\; \textcolor[rgb]{0.5,0.25,0.6}{return};\\
\quad\; \}\\
\quad\; newplot.synchronizeWith(synchronizer);\\
\quad\; \textcolor[rgb]{0.5,0.25,0.6}{super}.setSVGDocument(\\
\quad\; \quad\; newplot.getDocument());\\
\quad\; \textcolor[rgb]{0.5,0.25,0.6}{super}.setDisableInteractions(\\
\quad\; \quad\; newplot.getDisableInteractions());\\ \}\end{tabular}\\ \hline
Summ. & Attach to a new plot and display.\\ \hline\hline
\begin{tabular}[c]{@{}l@{}}Code\\ (Python)\end{tabular} &
\begin{tabular}[c]{@{}l@{}}\textcolor[rgb]{0.5,0.25,0.6}{def} \textcolor{blue}{get\_change\_lines\_in\_file\_for\_tag}(tag,\\
\quad\; \quad\; \quad\; \quad\; \quad\; \quad\; \quad\; \quad\; \quad\; \; change\_dict):\\
\quad\; cleaned\_lines = {[}{]}\\
\quad\; data\_list = change\_dict.get(\textcolor[rgb]{0.8,0.25,0.25}{'data'}, {[}{]})\\
\quad\; \textcolor[rgb]{0.5,0.25,0.6}{for} data\_dict \textcolor[rgb]{0.5,0.25,0.6}{in} data\_list:\\
\quad\; \quad\; block = data\_dict.get(\textcolor[rgb]{0.8,0.25,0.25}{'block'}, \textcolor[rgb]{0.8,0.25,0.25}{''})\\
\quad\; \quad\; lines = block.split(\textcolor[rgb]{0.8,0.25,0.25}{'\textbackslash{}\textbackslash{}n'})\\
\quad\; \quad\; \quad\; \textcolor[rgb]{0.5,0.25,0.6}{for} line \textcolor[rgb]{0.5,0.25,0.6}{in} lines:\\
\quad\; \quad\; \quad\; \quad\; index = line.find(tag)\\
\quad\; \quad\; \quad\; \quad\; \textcolor[rgb]{0.5,0.25,0.6}{if} (index \textgreater (-1)):\\
\quad\; \quad\; \quad\; \quad\; \quad\; line = line{[}index:{]}\\
\quad\; \quad\; \quad\; \quad\; \quad\; cleaned\_lines.append(line)\\
\quad\; \textcolor[rgb]{0.5,0.25,0.6}{return} cleaned\_lines\end{tabular}\\ \hline
Summ. & \begin{tabular}[c]{@{}l@{}}The received change\_dict is the jsonified version of\\ the changes to a file in a changeset being pushed to\\ the Tool Shed from the command line. This method\\ cleans and returns appropriate lines for inspection.\end{tabular}\\ \hline\hline
\end{tabular}%
}
\caption{Task samples of code summarization, where summ. refers to the output summary.}
\label{sample}
\end{table}

By 2020, software development and maintenance become an indispensable part of human work and life. Various assistant technical measures have been taken to facilitate more enjoyable software development, among which it is especially welcomed by programmers when there is a code summarization task generating sensible human language annotations automatically.

In early days, code summarization was a derivative problem of information retrieval \citep{haiduc2010use, eddy2013evaluating, wong2013autocomment, wong2015clocom} by matching the most similar code snippets which are labeled with summaries. Such method lacks generalization and performs unsatisfactorily. Thus in recent years, researchers treated code summarization as a task of language generation \citep{DBLP:conf/acl/IyerKCZ16, DBLP:conf/aaai/LiangZ18}, which usually depends on RNN-based Seq2Seq models \citep{DBLP:conf/emnlp/ChoMGBBSB14, DBLP:journals/corr/BahdanauCB14}.

It is already known that RNN-based models may encounter bottleneck when modeling long sequences due to its poor long-term dependency. For instance, a normal snippet of Java as shown in Table \ref{sample} usually has hundreds of tokens. More recently, \citet{ahmad-etal-2020-transformer} used an enhanced Transformer-based model to capture long-term and non-sequential information of source code, which outperformed previous RNN-based models by a large margin.

\begin{table}[t]
\centering
\resizebox{0.45\textwidth}{!}{%
\begin{tabular}{@{}lccc@{}}
\hline\hline
                  & \begin{tabular}[c]{@{}c@{}}Structure-\\ sensitive\end{tabular} & \begin{tabular}[c]{@{}c@{}}Long-term\\ dependency\end{tabular} & \begin{tabular}[c]{@{}c@{}}Feat-model\\ match\end{tabular} \\ \midrule
LSTM              &                                                                &                                                                &                                                            \\
Tree-LSTM         & \checkmark                                                              &                                                                &                                                            \\
Transformer       &                                                                & \checkmark                                                              &                                                            \\
LSTM + SBT        & \checkmark                                                              &                                                                & \checkmark                                                          \\
Transformer + SBT & \checkmark                                                              & \checkmark                                                              &                                                            \\
SiT              & \checkmark                                                              & \checkmark                                                              & \checkmark                                                          \\ \hline\hline
\end{tabular}%
}
\caption{Comparison of the previous models with proposed SiT model. The last column refers to whether input features match with the corresponding model.}
\label{t1}
\end{table}

On the other hand, in the light of the structural nature of programming languages, structure clues are supposed to greatly enhance programming language processing task like code summarization \citep{DBLP:conf/iclr/FernandesAB19}. Indeed, substantial empirical studies showed that Abstract Syntax Tree may help models better comprehend code snippets and achieve more sensible generation results. Previous approaches could be divided into two categories. The first is to employ non-sequential encoders (e.g., TBCNN \citep{DBLP:conf/aaai/MouLZWJ16}, Tree-LSTM \citep{shido2019automatic}, Tree-Transformer \citep{harer2019tree}, Graph Neural Network \citep{allamanis2018learning, liu2020automatic, leclair2020codegnn, wang2021learning}) to directly model structural inputs. The other is to pre-process structural inputs to apply sequential models on them. \citet{alon2018codeseq} used LSTM to encode code structure by sampling possible paths of AST. Another similar work is structure-based traversal (SBT) \citep{hu2018deep}, which manages to flatten ASTs into linear sequences.

Though existing studies achieve success on the concerned code summarization task more or less, there is still room in improving both of the above modeling approaches. It is well known RNN encoders like LSTM only have limited capabilities in capturing long-range dependencies in sequence, and GNN-like models may be too sensitive to local information, which casts a natural solution, what if incorporating SBT into the Transformer? However, it is surprising that SBT only works effectively with LSTM but not the Transformer according to \citet{ahmad-etal-2020-transformer}. We attribute this to the linear and nonlinear inconsistence between SBT and encoder forms. SBT enables sequential encoders to learn non-sequential relationship (such as syntax) still in a certain elaborate linear forms. RNN may be effectively enhanced by SBT right because of its sequential architecture through attention mechanism. Transformer learns features through self-attention network (SAN), nevertheless which acts more like a non-sequential process. Consequently, such sequential features are unsuitable for a non-sequential architecture to extract implicit structural information. We boldly call it Feature-Model Match problem in Table \ref{t1}. In this paper, we thus design an improved Transformer variants, structure-induced Transformer (SiT) to alleviate such difficulty in terms of a structure-induced self-attention mechanism, so that the resulted model may enjoy both merits, capturing long-range dependencies and more global information. The proposed model design has been applied to benchmark datasets and helps achieve new state-of-the-art performance.

\begin{figure}[ht]
\centering
\includegraphics[width=0.45\textwidth]{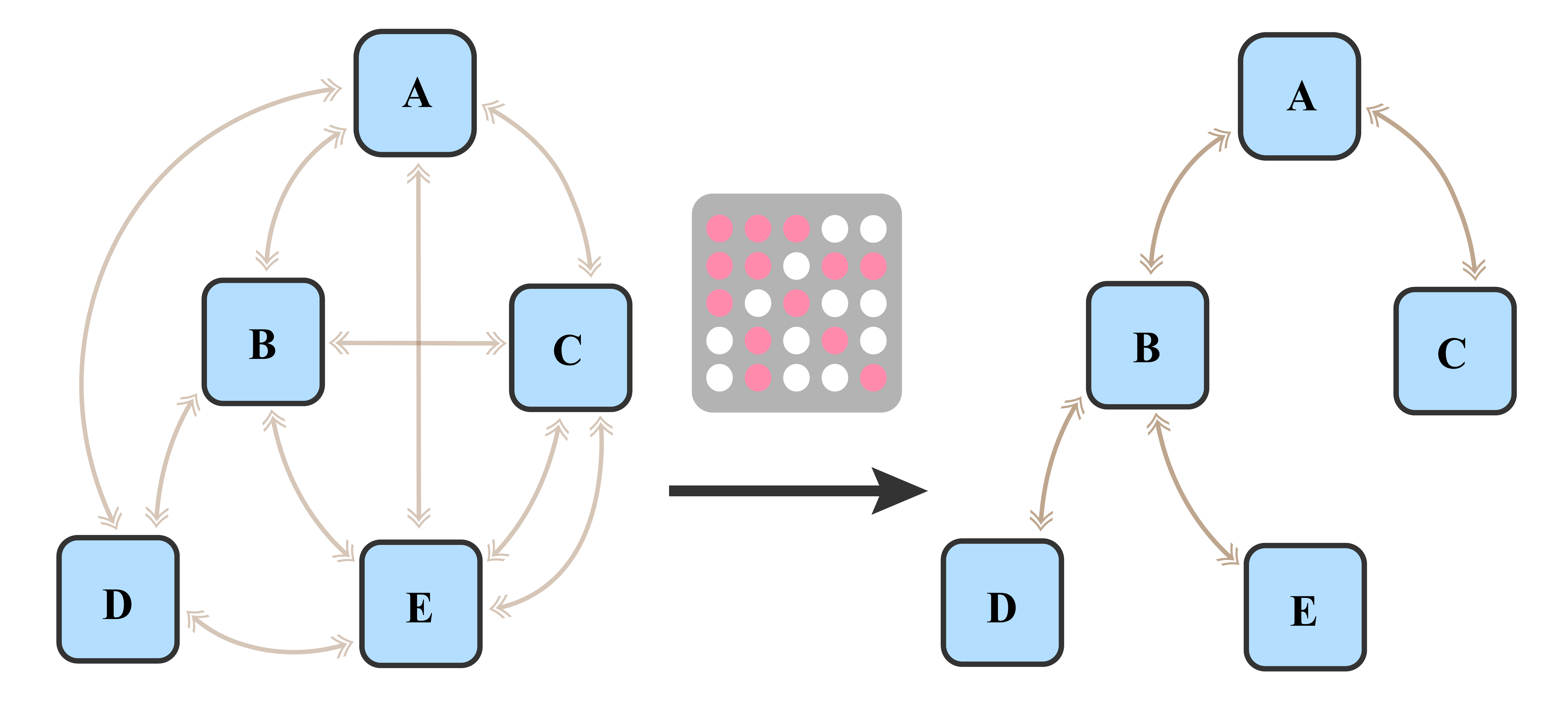}
\caption{Use of adjacency matrix to transform original self-attention, left-hand complete graph, into structure-induced self-attention, right-hand graph which looks clear-cut. Note that we omit self-circles for concision.}
\label{2}
\end{figure}

\section{Structure-based Code Summarization}

The following sections present our code summarization method with two parts, in which the first is about structure representation of code, and the second is our proposed structure-induced Transformer.

\subsection{Structure Representation of Code}

Note that programming language like source code is subtle that certain different formats may result in different compilations. Thus pre-processing could be an great impact in code summarization.

\begin{figure}[ht]
\centering
\includegraphics[width=0.45\textwidth]{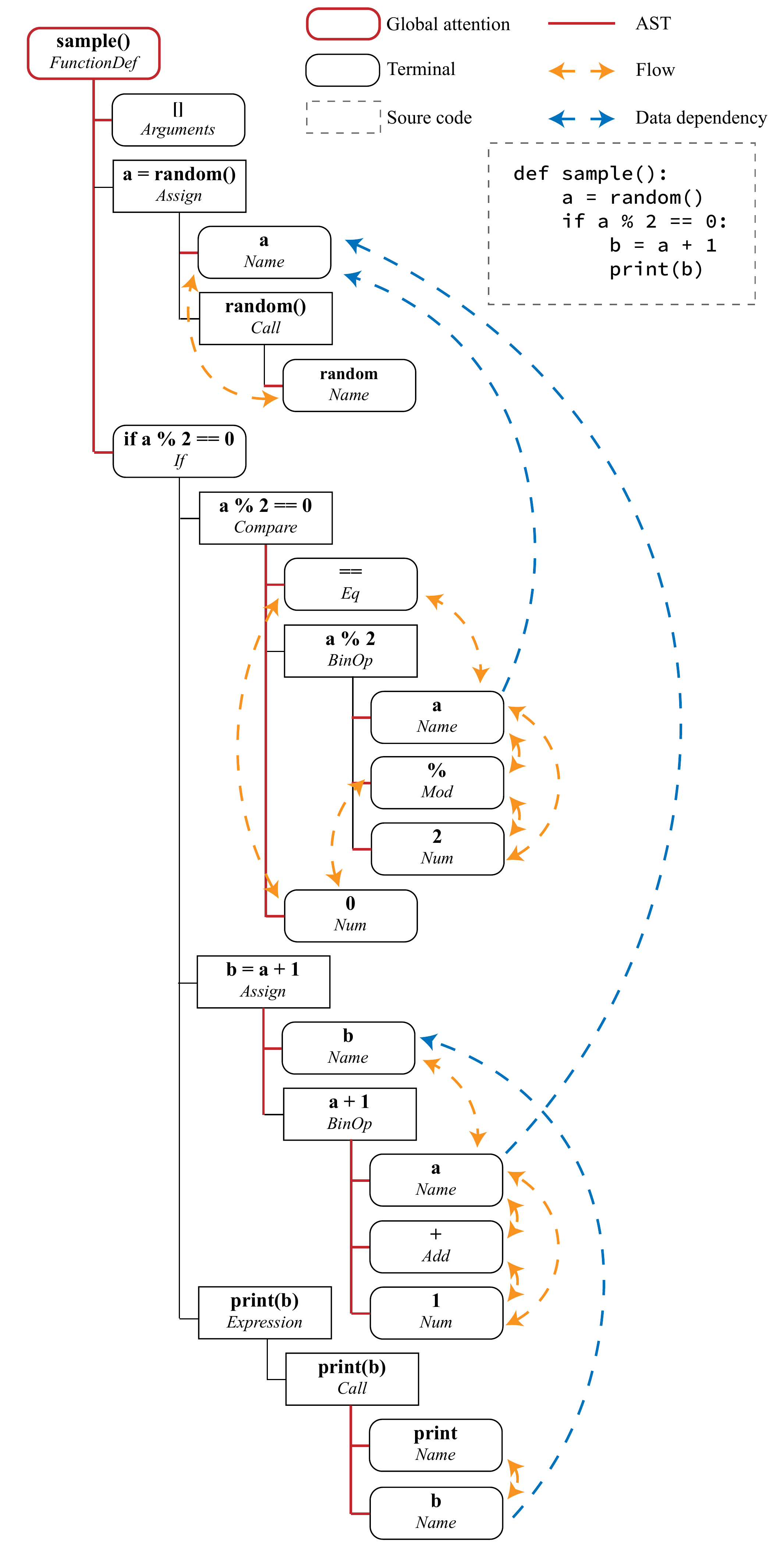}
\caption{A Python code sample of multi-view graph used in Si-SAN. The code snippet is referred from \citet{liu2020automatic}, which is original in Java.}
\label{8}
\end{figure}

We adopt Abstract Syntax Tree (AST) for representing the language grammar of source code as usual. Figure \ref{8} depicts a typical AST, which is composed of terminal nodes and non-terminal nodes. A non-terminal node represents certain construction like \emph{If} and \emph{BinaryOp}, while terminal nodes represent its semantic components, such as identifiers and numbers.

In model implementation, we adopt adjacency matrix $ A $ to represent the AST instead of structure based traversal method as in \citet{hu2018deep}, which represents tree structure in a sequential format. Such choice is well compatible with Transformer, which calculates attention weights by performing a dot-product of key-query pairs and results in an attention matrix of $ l \times l $. We let $ l $ equal to number of AST nodes, then code summarization with Transformer becomes possible through applying a position-wise multiplication of $ A $ and original attention matrix.

Inspired by Code Property Graph (CPG) \citep{DBLP:conf/sp/YamaguchiGAR14, liu2020automatic}, we further expand AST into a multi-view network (MVN or multi-view graph) \citep{sindhwani2005co, DBLP:conf/icml/ZhouB07, DBLP:conf/nips/KumarRD11}. An MVN is composed of multiple views, each view corresponding to a type of structural relationships while all views sharing the same set of vertexes \citep{DBLP:journals/corr/abs-1801-06597}. In this paper, we construct a three-view graph based on different code semantics, which are abstract syntax, control flow and data dependency. We show an example in Figure \ref{8}, where we use colorful strokes to describe different compositions in the graph. Note that we only utilize terminal nodes which are marked as rounded rectangles.

Specifically, we first generate an AST, on the basis of which we add additional edges to further represent the flow of control and data. For control flow, since Transformer is order-sensitive with position encoding, we only need to focus on each statement node. For instance, nodes \emph{b}, \emph{=}, \emph{a}, \emph{+}, \emph{1} make a complete statement \emph{b=a+1}. We connect each of them since they are in the same execution order. For data dependencies, we connect relevant data across the whole program, as the variable \emph{b} in expression \emph{print(b)} and assignment \emph{b=a+1} respectively, where the former is defined and loaded from the latter.

Now we may obtain three adjacency matrices of syntax, flow and dependency respectively, which are colored in red, yellow and blue in Figure \ref{8}. We combine them together and finally obtain a multi-view graph. Additionally, we add global attention on the root, which is allowed to attend to all tokens in the code, and all tokens in the code can attend to it. With aggregated structure, our structure-based code summarization is expected to capture various semantics of programs.

Note that our multi-view graph is different from CPG. which is original for C/C++ only and we do not find an appropriate analysis platform for other languages.

\begin{figure*}
\centering
\includegraphics[width=0.95\textwidth]{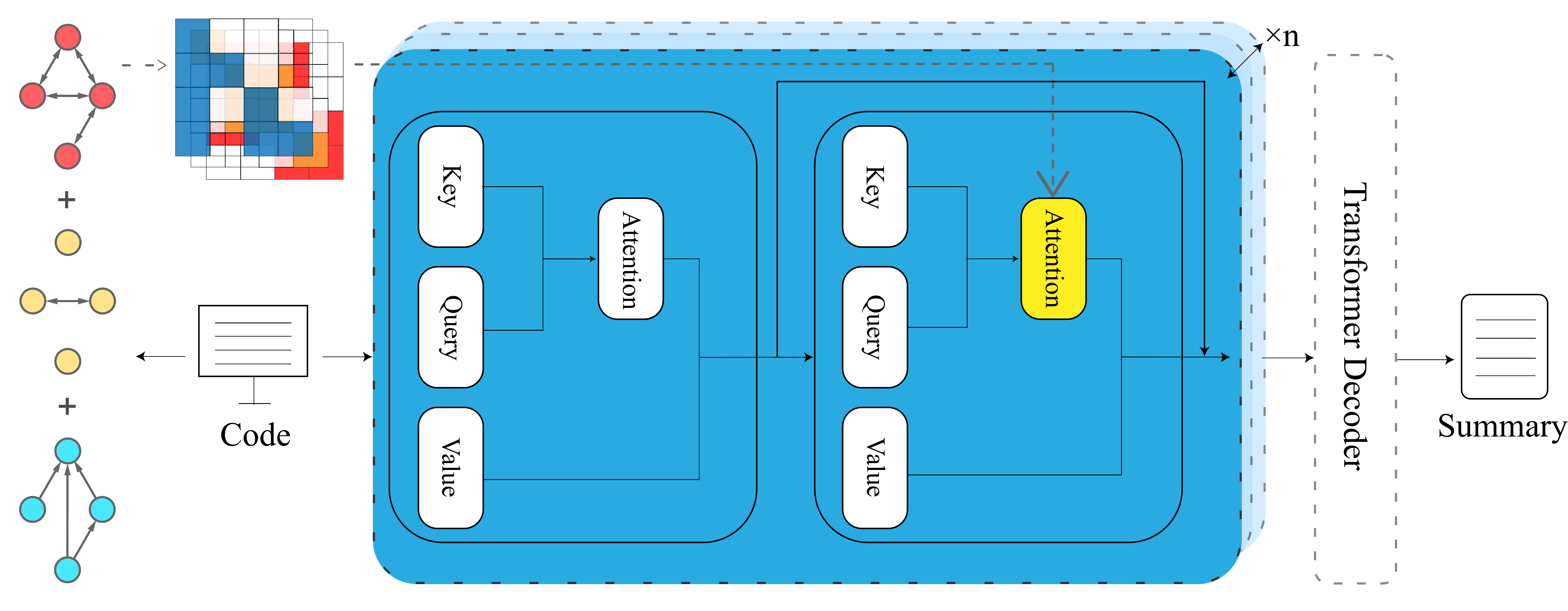}
\caption{Overall architecture of Structure-induced Transformer (SiT).}
\label{3}
\end{figure*}

\subsection{Structure-induced Transformer}

Followed by appropriate structure representation and graph construction, we now propose our structure-induced Transformer (SiT) for code summarization, which is a structure-sensitive transformer \citep{DBLP:conf/aaai/0001WZDZ020, DBLP:conf/emnlp/NarayanMAPBM20, xu-etal-2020-discourse} model and is able to comprehend code snippets both semantically and syntactically. Meanwhile, we do not introduce extra parameters in SiT so that guarantee the training efficiency. In this section, we first review the self-attention network (SAN) of Transformer in terms of attention graph. Then we correspondingly propose structure-induced self-attention to build the structure-induced Transformer.

\paragraph{Vanilla Self-Attention} Transformer is composed of stacks of identical layers for both encoder and decoder \citep{vaswani2017attention}. Each layer emphasizes on self-attention mechanism, which is denoted as:
\begin{equation}
\label{e1}
SAN(X) = Softmax\left(\frac{Q K^{T}}{\sqrt{d_{k}}}\right)V
\end{equation}
where $ X=(x_{1},\ldots,x_{l}) $ denotes the input sequence of sub-words, $ l $ denotes the sequence length and $ d_{k} $ denotes the hidden size per head. Now we view each sub-word as a vertex $ n $ and inner product of each key-value pair as a directed edge $ e $, the SAN can be described as a directed cyclic graph. Equation \ref{e1} can be rewritten as follow:
\begin{equation}
\label{e2}
SAN(X)=E \cdot N
\end{equation}
The attention scores $ E=\{e_{ij}\} $ refers to a weight matrix of edges where $ e_{ij} $ represents how significant node $ n_{i} $ attend to node $ n_{j} $, while value matrix $ N=\{n_i\} $ refers to each node representation. Figure \ref{2} depicts the process of calculating attention scores.

Note that SAN actually generates a fully connected cyclic graph without consideration of the very needed structure-aware representation for our concerned task.

\paragraph{Structure-induced Self-Attention} To represent the needed structure information, we propose structure-induced self-attention network (Si-SAN).

Specifically, we introduce multi-view network into Equation \ref{e1}, that is, multiply the adjacency matrix by key-query pairs:
\begin{equation}
\label{e3}
SiSAN(X) = Softmax\left(\frac{A_{mv} \cdot Q K^{T}}{\sqrt{d_{k}}}\right)V
\end{equation}
where $ A_{mv} $ refers to the multi-view representation of code.

Note that Si-SAN does not change the input code but appropriately incorporate code structure into SAN by changing its attention pattern. As shown in Figure \ref{2}, when $ a_{ij}=0 $ in $ A_{mv} $, the attention between $ n_{i} $ and $ n_{j} $ will be dropped out \citep{DBLP:journals/corr/abs-2104-04692}. We consequently obtain a more explicit attention graph. Different from calculating global information onto the whole sentence in original SAN, Si-SAN is expected to calculate structural information more accurately.

\begin{table*}[t]
\centering
\resizebox{\textwidth}{!}{%
\begin{tabular}{@{}l|lll|lll@{}}
\hline\hline
\multirow{2}{*}{\textbf{Model}} & \multicolumn{3}{c|}{\textbf{Java}} & \multicolumn{3}{c}{\textbf{Python}} \\ \cline{2-7}
& \multicolumn{1}{l}{\textbf{BLEU}} & \multicolumn{1}{l}{\textbf{ROUGE-L}} & \multicolumn{1}{l|}{\textbf{METEOR}} & \multicolumn{1}{l}{\textbf{BLEU}} & \multicolumn{1}{l}{\textbf{ROUGE-L}} & \multicolumn{1}{l}{\textbf{METEOR}} \\ \hline
CODE-NN \citep{DBLP:conf/acl/IyerKCZ16} & \multicolumn{1}{l}{27.60}         & \multicolumn{1}{l}{41.10}            & \multicolumn{1}{l|}{12.61}           & \multicolumn{1}{l}{17.36}         & \multicolumn{1}{l}{37.81}            & \multicolumn{1}{l}{09.29}           \\
Tree2Seq \citep{DBLP:conf/acl/EriguchiHT16} & \multicolumn{1}{l}{37.88}         & \multicolumn{1}{l}{51.50}            & \multicolumn{1}{l|}{22.55}           & \multicolumn{1}{l}{20.07}         & \multicolumn{1}{l}{35.64}            & \multicolumn{1}{l}{08.96}           \\
Hybrid2Seq \citep{wan2018improving} & \multicolumn{1}{l}{38.22}         & \multicolumn{1}{l}{51.91}            & \multicolumn{1}{l|}{22.75}           & \multicolumn{1}{l}{19.28}         & \multicolumn{1}{l}{39.34}            & \multicolumn{1}{l}{09.75}           \\
DeepCom \citep{hu2018deep} & \multicolumn{1}{l}{39.75}         & \multicolumn{1}{l}{52.67}            & \multicolumn{1}{l|}{23.06}           & \multicolumn{1}{l}{20.78}         & \multicolumn{1}{l}{37.35}            & \multicolumn{1}{l}{09.98}           \\
API + Code \citep{ijcai2018-314} & \multicolumn{1}{l}{41.31}         & \multicolumn{1}{l}{52.25}            & \multicolumn{1}{l|}{23.73}           & \multicolumn{1}{l}{15.36}         & \multicolumn{1}{l}{33.65}            & \multicolumn{1}{l}{08.57}           \\
Dual Model \citep{wei2019code} & \multicolumn{1}{l}{42.39}         & \multicolumn{1}{l}{53.61}            & \multicolumn{1}{l|}{25.77}           & \multicolumn{1}{l}{21.80}         & \multicolumn{1}{l}{39.45}            & \multicolumn{1}{l}{11.14}           \\
Transformer \citep{ahmad-etal-2020-transformer} & \multicolumn{1}{l}{44.58}         & \multicolumn{1}{l}{54.76}            & \multicolumn{1}{l|}{26.43}           & \multicolumn{1}{l}{32.52}         & \multicolumn{1}{l}{46.73}            & \multicolumn{1}{l}{19.77}           \\ \hline\hline
Transformer$^\ast$ \citep{ahmad-etal-2020-transformer} & \multicolumn{1}{l}{44.87}         & \multicolumn{1}{l}{54.95}            & \multicolumn{1}{l|}{26.58}           & \multicolumn{1}{l}{32.85}         & \multicolumn{1}{l}{46.93}            & \multicolumn{1}{l}{19.86}           \\
SiT
& \multicolumn{1}{l}{\textbf{45.76}($\uparrow$1.18)}
& \multicolumn{1}{l}{\textbf{55.58}($\uparrow$0.82)}
& \multicolumn{1}{l|}{\textbf{27.58}($\uparrow$1.15)}
& \multicolumn{1}{l}{\textbf{34.11}($\uparrow$1.59)}
& \multicolumn{1}{l}{\textbf{48.35}($\uparrow$1.62)}
& \multicolumn{1}{l}{\textbf{21.11}($\uparrow$1.34)}           \\
CodeBERT$^\ast\dag$ \citep{feng2020codebert} & \multicolumn{1}{l}{43.33}         & \multicolumn{1}{l}{54.64}            & \multicolumn{1}{l|}{26.20}           & \multicolumn{1}{l}{33.47}         & \multicolumn{1}{l}{49.35}            & \multicolumn{1}{l}{21.69}           \\
SiT on CodeBERT$\dag$
& \multicolumn{1}{l}{\textbf{45.19}($\uparrow$0.61)}
& \multicolumn{1}{l}{\textbf{55.87}($\uparrow$1.11)}
& \multicolumn{1}{l|}{\textbf{27.52}($\uparrow$1.09)}
& \multicolumn{1}{l}{\textbf{34.31}($\uparrow$1.79)}
& \multicolumn{1}{l}{\textbf{49.71}($\uparrow$2.98)}
& \multicolumn{1}{l}{\textbf{22.09}($\uparrow$2.32)}           \\ \hline\hline
\end{tabular}%
}
\caption{BLEU, ROUGE-L and METEOR for our approach compared with other baselines. $\dag$ refers to pre-trained models while $^\ast$ refers to models we rerun. The results of upper part are directly reported from \citet{ahmad-etal-2020-transformer}. Note that we only rerun Transformer and CodeBERT since they are much stronger than the other baselines. However, our results are even stronger. We show the ranges compared to the Transformer in \citet{ahmad-etal-2020-transformer}.}
\label{t2}
\end{table*}
\begin{figure*}[]
\centering
\subfigure[BLEU score with training steps on Java]{
\includegraphics[width=0.48\textwidth]{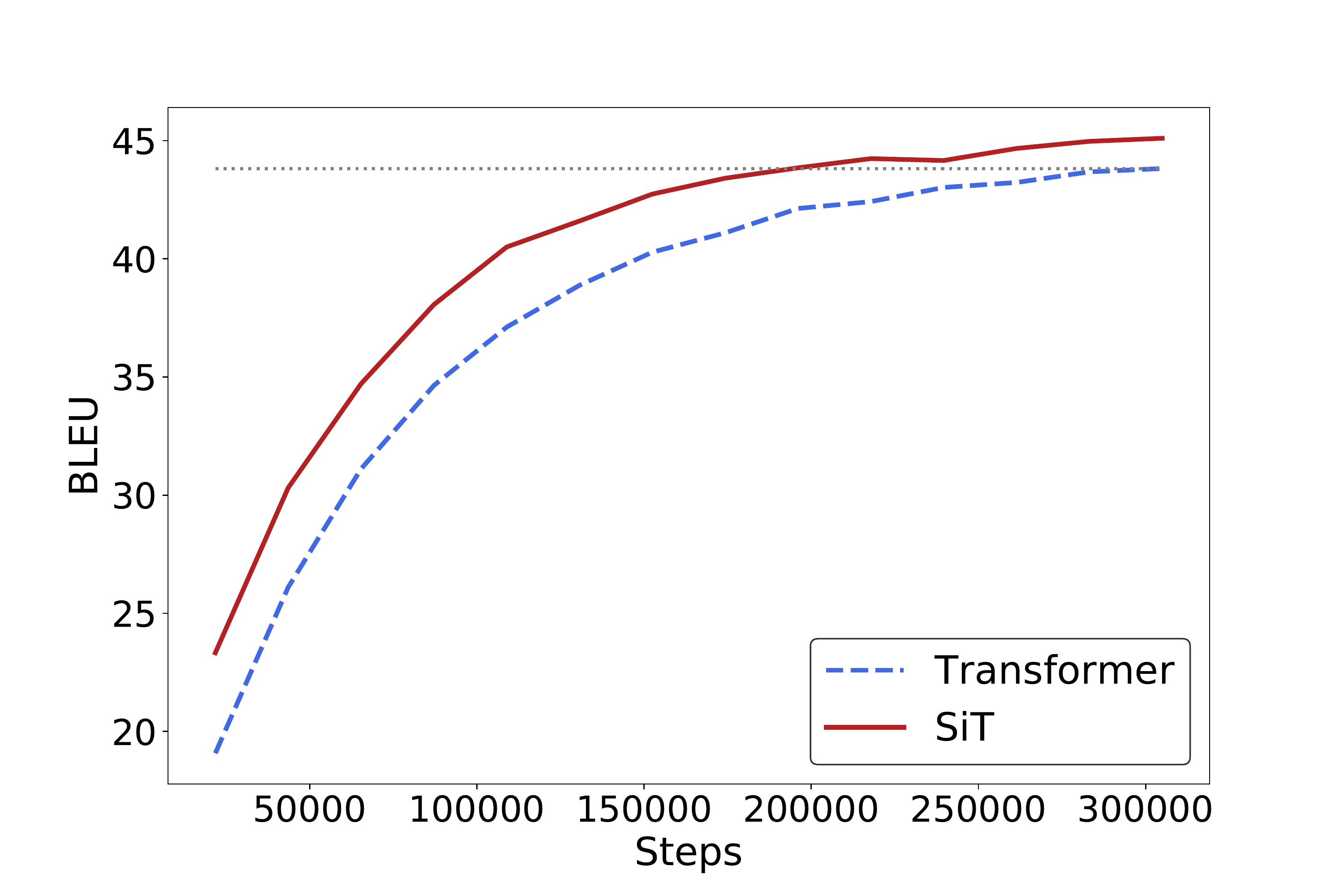}
}
\subfigure[BLEU score with training steps on Python]{
\includegraphics[width=0.48\textwidth]{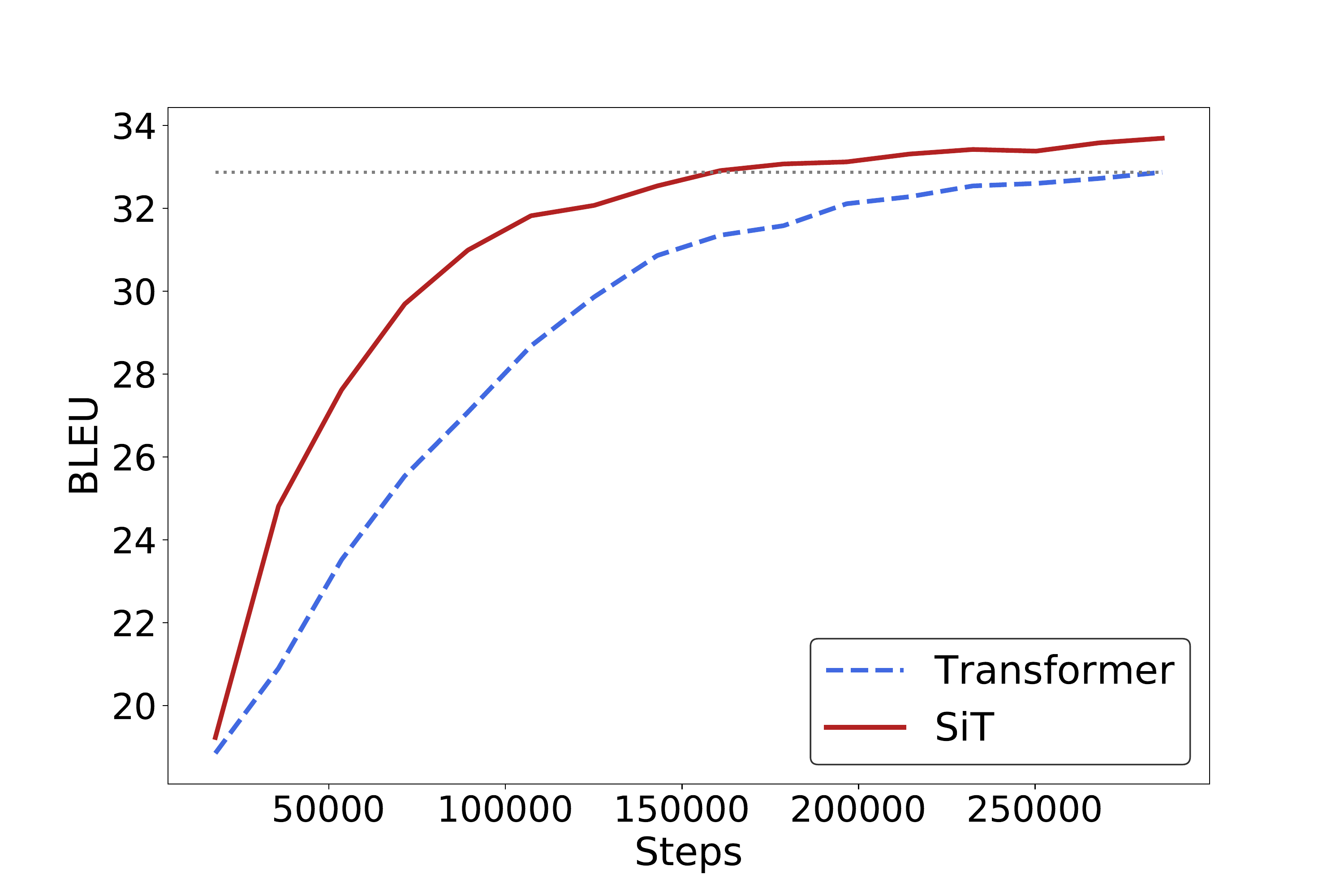}
}
\caption{Convergence between Transformer and SiT.}
\label{5}
\end{figure*}

\paragraph{Structure-induced Module} To enhance robustness and avoid over-pruning, we introduce structure-induced module, which is a stack of two layers, SAN and Si-SAN. In each module, SAN is followed by Si-SAN and the output is the combination of both layers. Specifically, given input sequence $ X=(x_{1},\ldots,x_{l}) $, where $ l $ denotes sequence length, we first pass it through an SAN layer to obtain hidden representation denoted as $ H=(h_{1},\ldots,h_{l}) $:
\begin{equation}
\label{e4}
H = Concat(SAN_{1}(X),\ldots,SAN_{h}(X))
\end{equation}
where $ h $ refers to number of heads of multi-head attention while $ SAN_{i} $ refers to self-attention of head $ i $. Subsequently, we pass $ H $ through a Si-SAN layer to obtain $ H'=(h'_{1},\ldots,h'_{l}) $:
\begin{equation}
\label{e5}
H' = Concat(SiSAN_{1}(H),\ldots,SiSAN_{h}(H))
\end{equation}
Finally, we use an aggregation to fuse $ H $ and $ H' $ to obtain final representation $ \bar{H}=(\bar{h_{1}},\ldots,\bar{h_{l}}) $:
\begin{equation}
\label{e6}
\bar{H} = Aggr(H,H')
\end{equation}
where the aggregation we use is simple position-wise sum. We explore that the structure-induced module is more robust and leads to a better performance. In each stack, model begins to learn global information with SAN, where all connections are available. Subsequently, through Si-SAN, model is told which of the connections are useful and which should be shut down and thus avoiding over-pruning. Note that SiT with 3 stacks of structure-induced modules still consists of 6 encoder layers and 6 decoder layers, but only changes the architecture between modules of Transformer, not introducing any extra parameters.

Figure \ref{3} depicts the overall architecture of SiT. Compared to original Transformer, our SiT with Si-SAN encodes a more accurate relative representation of code through pruning redundant connections.

\subsection{SiT-based Code Summarization}

Based on our structure-induced Transformer (SiT), now we specify our code summarization process. 

We first transform the input code into adjacency matrices of multiple views and combine them through a weighted sum:
\begin{equation}
\label{e7}
A_{mv}=\alpha A_{ast} + \beta A_{fl} + \gamma A_{dp}
\end{equation}
where $ \alpha,\beta,\gamma $ refer to the corresponding weight for each view. Then we pass code sequences and corresponding adjacency matrices into SiT encoder, which contains 3 Si-SAN layers. For decoder, we apply original Transformer decoder with cross attention. Finally, the summarization of the input code is generated through autoregressive decoding.

\section{Experiments}

\subsection{Datasets and Pre-processing}

\paragraph{Datasets} Our experiments are conducted on two benchmarks of Java \citep{hu2018deep} and Python \citep{wan2018improving}, and for both we follow their training, test and development divisions.
\paragraph{Graph Construction} For Java code, we refer to the method provided in \citep{hu2018deep}. They use \emph{javalang} module of Python to compile Java and fetch AST in a dictionary form. For Python code, we generate trees by ourselves based on \emph{ast} and \emph{asttokens} modules. Finally, we write a script to resolve ASTs into multi-view adjacency matrices\footnote{https://github.com/gingasan/astruc}, where we let $ \alpha=\beta=\gamma=1 $ for all experiments\footnote{We try to adjust the weights of three views, showing little performance variant, which suggests that self-attention network itself may balance the relative significance between the three.}.
\paragraph{Out-Of-Vocabulary} Code corpus in programming language may have a much bigger vocabulary than natural language, including vast operators and identifiers. We have to introduce vast out-of-vocabulary (OOV) tokens (usually replaced by $ \langle$UNK$\rangle $) \citep{hu2018deep} to keep it in a regular size. To avoid OVV problem, we apply \emph{CamelCase} and \emph{snake\_case} tokenizers \citep{ahmad-etal-2020-transformer} to reduce code vocabulary and remove all extra nodes which do not correspond to specific tokens.

\subsection{Baselines}

We take all three categories of state-of-the-art models as our baselines for comparison.
\paragraph{Transformer} We refer to the enhanced Transformer in \citep{ahmad-etal-2020-transformer} which equipped with copy attention \citep{DBLP:conf/acl/SeeLM17} and relative position encoding (RPE) \citep{shaw-etal-2018-self}. For fair enough comparison, we run their model on our machine under the same environment with SiT. Note that we also utilize RPE in SiT because of its better capability in capturing long sequences, while we do not utilize copy attention.
\paragraph{LSTM} This group includes all relevant LSTM models with sequential and non-sequential inputs \citep{DBLP:conf/acl/IyerKCZ16, DBLP:conf/acl/EriguchiHT16, wan2018improving, hu2018deep, ijcai2018-314, wei2019code}.
\paragraph{Pre-trained Language Model} We also compare our model with CodeBERT \citep{feng2020codebert}, a pre-trained language model on both natural and programming languages. It is pre-trained over six programming languages with MLM \citep{DBLP:conf/naacl/DevlinCLT19} and RTD \citep{DBLP:conf/iclr/ClarkLLM20}.

\subsection{Training Details}

We train our model on a single nVidia Titan RTX with batch size in \{32, 64\}. The learning rate is in \{3e-5, 5e-5\} with warm-up rate of 0.06 and L2 weight decay of 0.01. The maximum number of epochs is set to 150 for Transformer and 30 for CodeBERT. For validation, we simply use greedy search, while for evaluation, we use beam search with beam size in \{4, 5, 8\} and choose the best result\footnote{https://github.com/gingasan/sit3}.

\begin{figure}[t]
\centering
\subfigure[Full]{
\includegraphics[width=0.22\textwidth]{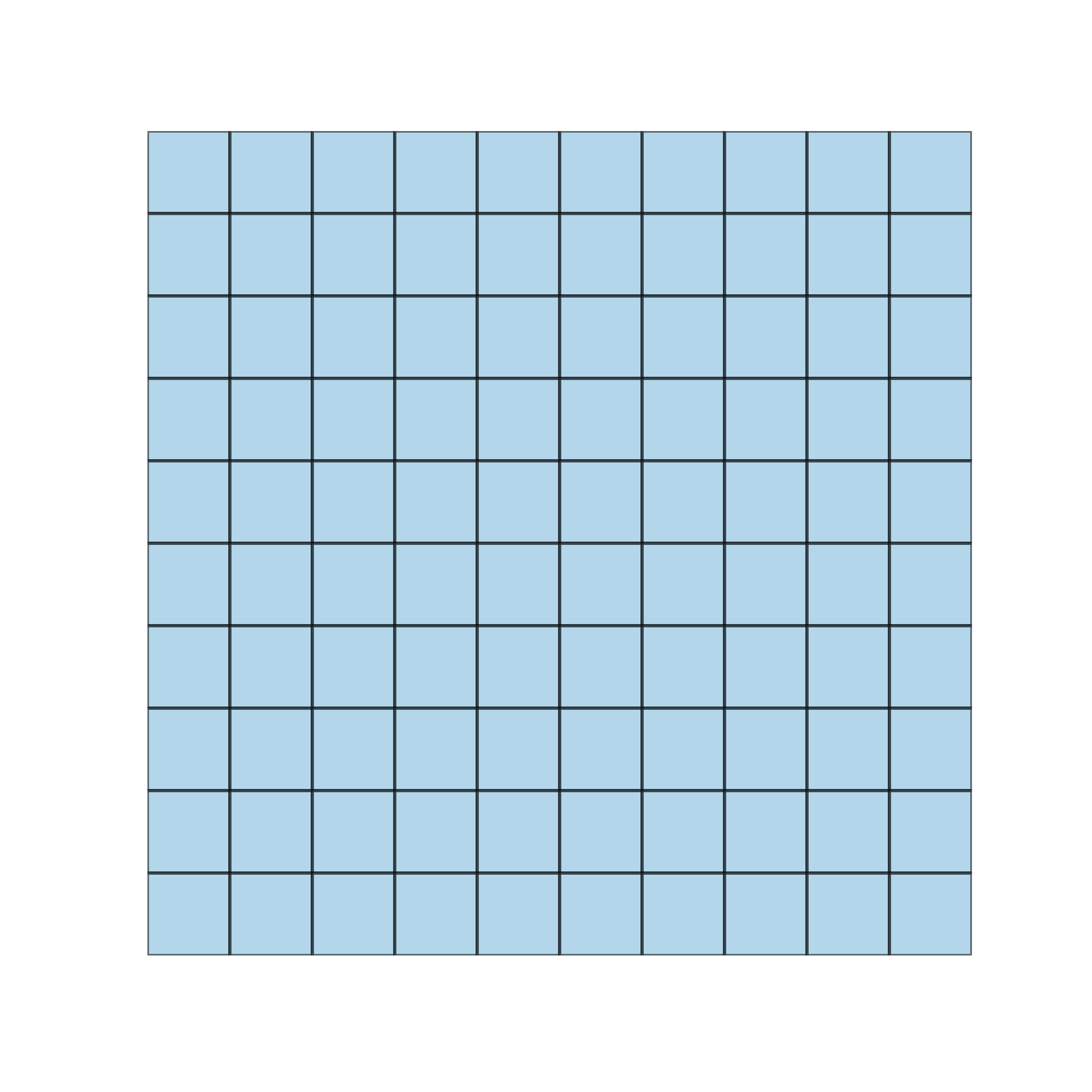}
}
\subfigure[Window]{
\includegraphics[width=0.22\textwidth]{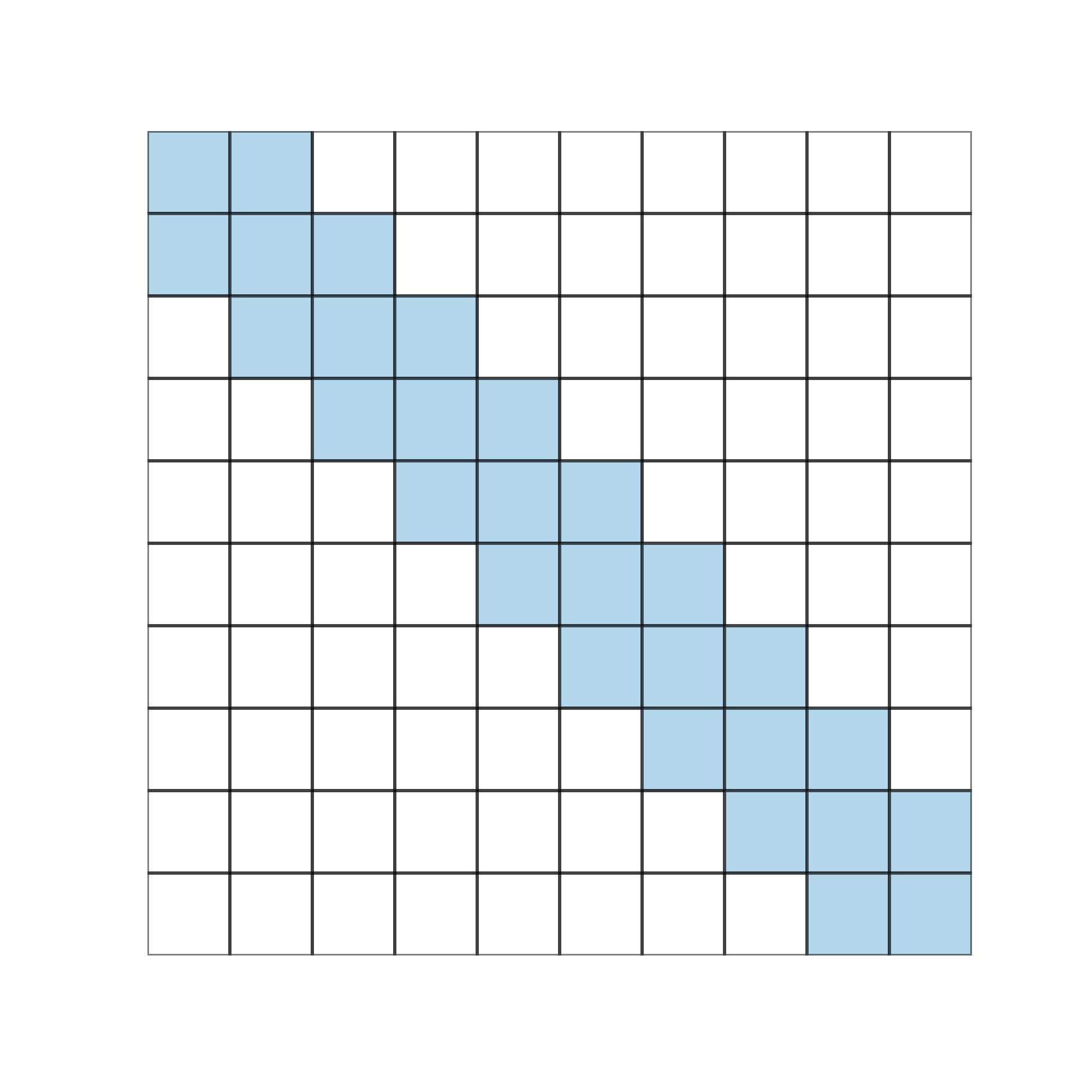}
}

\subfigure[Random]{
\includegraphics[width=0.22\textwidth]{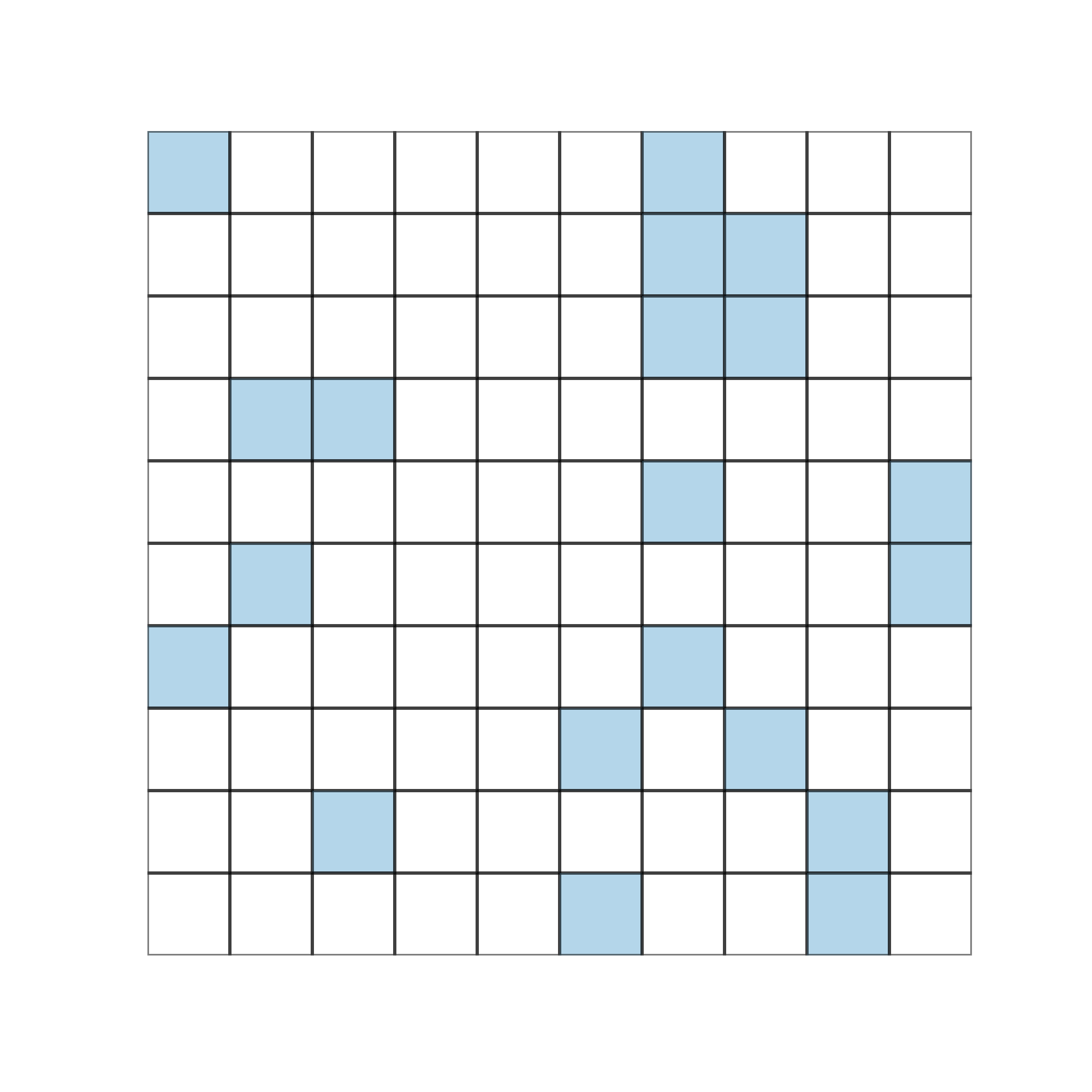}
}
\subfigure[Structure-induced]{
\includegraphics[width=0.22\textwidth]{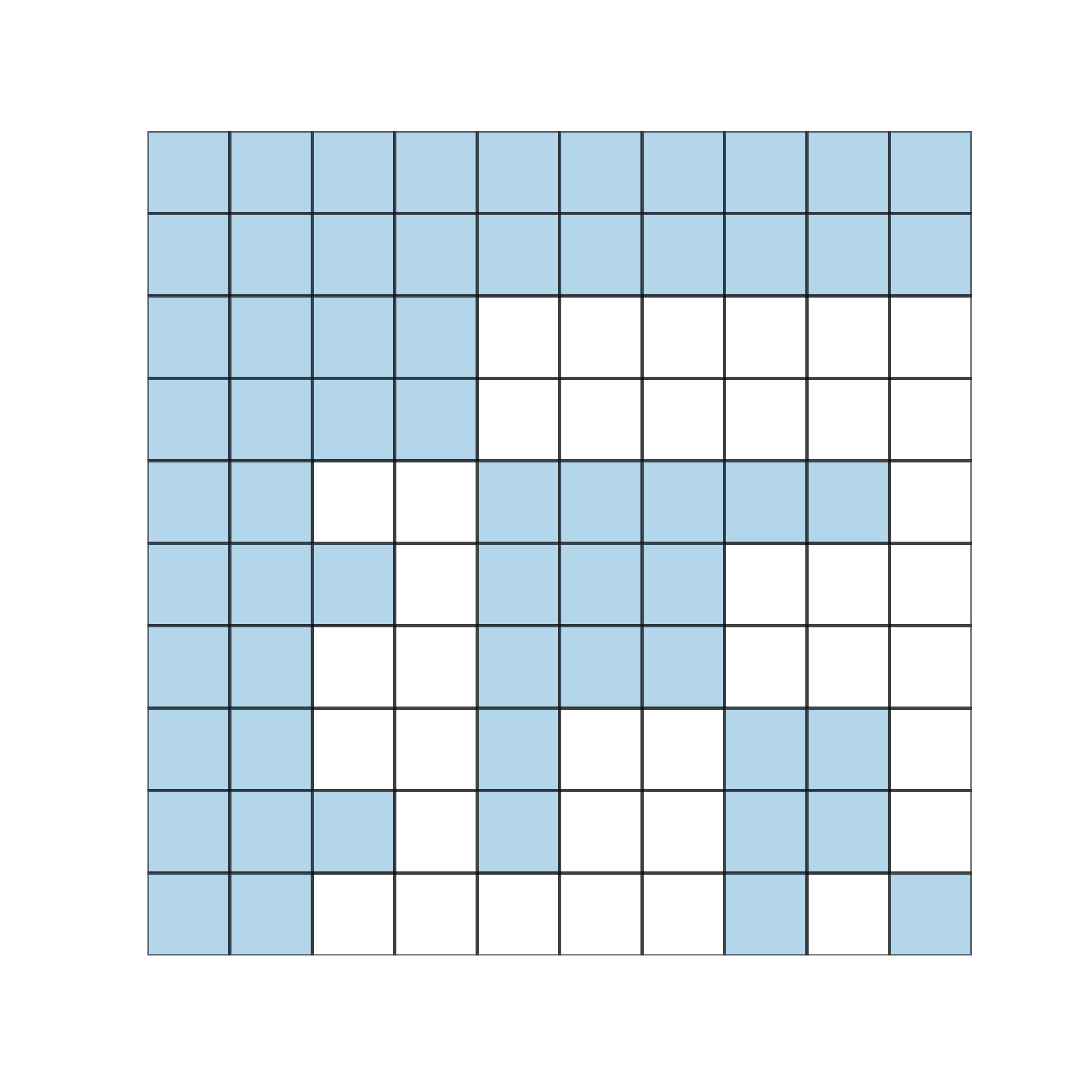}
}
\caption{Comparison of different types of self-attention pattern. (b) Window attention with $ w=2 $. (c) Random attention with $ r=2 $.}
\label{4}
\end{figure}

\subsection{Main Results}

\paragraph{Scores} Table \ref{t2} shows the overall results on Java and Python benchmarks. The Transformer baseline is strong enough as it outperforms all the previous works by a significant margin. However, our model is more powerful, further boosting Transformer with more than 1 BLEU points on Java and Python respectively and achieves new state-of-the-art results. Specifically, SiT achieves higher scores on Python, increasing by 1.59, 1.62 and 1.34 points on BLEU, ROUGE-L and METEOR respectively. According to dataset statistics, Python contains 5 times more unique code tokens than Java, which makes it much more challenging. Thus the superiority of SiT on Python tends to be notable. Even so, SiT still boosts Transformer by 1.18, 0.82 and 1.15 points on BLEU, ROUGE-L and METEOR respectively on Java.

\paragraph{Convergence} Moreover, Figure \ref{5} shows the trend of BLEU scores on development set over training steps. SiT achieves a much faster convergence rate than Transformer. For instance on Python dataset, SiT arrives the best performance of Transformer in about 100 epochs, while the latter one still needs 50 more to finally achieve the optimal. Note that the running time of each epoch for both models is the same. Such high convergence rate helps showcase the necessity of Si-SAN.

\paragraph{Pre-training} On the other hand, we can see that CodeBERT also achieves competitive results on both Java and Python. However, SiT is still more powerful on most metrics, which outperforms CoderBERT by 2.15, 0.95 and 1.15 points on BLEU, ROUGE-L and METEOR respectively on Java. However, CodeBERT performs much better on Python, which outperforms SiT by 1.00 and 0.58 points on ROUGE-L and METEOR. Note that CodeBERT is much bigger in size than Transformer and SiT (see Appendix \ref{b}).

For further verification, we follow CodeBERT and conduct a RoBERTa-based \citep{DBLP:journals/corr/abs-1907-11692} SiT to further fine-tune on both Java and Python. As shown in Table \ref{t2}, pre-trained SiT obtains attractive results, further improving CodeBERT on all the metrics, which implies that our elaborate encoder design is still effective even under powerful pre-training assistance.

\section{Ablation Study and Analysis}

This section reports our ablation studies to valid our model on the dataset of Python-V2\footnote{https://github.com/EdinburghNLP/code-docstring-corpus/tree/master/V2} \citep{DBLP:conf/ijcnlp/BaroneS17}, in which we conduct standard and unified pre-processing for strict fair comparison.

\begin{table}[]
\centering
\resizebox{0.46\textwidth}{!}{%
\begin{tabular}{@{}l|cccc@{}}
\hline\hline
\textbf{Model}   & \textbf{Prop.} & \textbf{BLEU} & \textbf{ROUGE-L}  & \textbf{METEOR} \\ \hline
Transformer & 0     & 47.42  & 57.28  & 29.62 \\
Transformer & 50\%  & 49.64  & 59.39  & 31.16 \\
Transformer & 100\% & 49.80  & 59.38  & 31.30 \\
SiT         & 50\%  & \textbf{50.04}  & \textbf{59.56}  & \textbf{31.46} \\ \hline\hline
\end{tabular}%
}
\caption{BLEU, ROUGE-L and METEOR for variant models with incremental proportions of Si-SAN.}
\label{t3}
\end{table}
\begin{table}[]
\centering
\resizebox{0.46\textwidth}{!}{%
\begin{tabular}{@{}l|cccc@{}}
\hline\hline
\textbf{Model} & \textbf{Attn.} & \textbf{BLEU} & \textbf{ROUGE-L} & \textbf{METEOR} \\ \hline
Transformer    & Full       & 47.42     & 57.28     & 29.62 \\  
Transformer    & Window     & 49.28     & 58.80     & 30.90 \\
Transformer    & Random     & 38.06     & 57.28     & 22.76 \\
Transformer    & Struc.     & \textbf{49.80}     & \textbf{59.38}     & \textbf{31.30} \\ \hline\hline
\end{tabular}%
}
\caption{BLEU, ROUGE-L and METEOR for variant models with different attention patterns.}
\label{t4}
\end{table}
\begin{table}[]
\centering
\resizebox{0.46\textwidth}{!}{%
\begin{tabular}{@{}l|cccc@{}}
\hline\hline
\textbf{Model}                                              & \textbf{BLEU} & \textbf{RE.-L} & \textbf{MTR.} & \textbf{SPEED} \\ \hline
Transformer        & 44.87         & 54.95        & 26.58           & 1.0x            \\
Transformer + SBT  & 43.34         & 53.97        & 25.02           & 1.5x            \\
SiT-AST only       & 45.43         & 55.30        & 27.21           & 1.0x            \\ 
SiT                & \textbf{45.76}         & \textbf{55.58}        & \textbf{27.58}           & \textbf{1.0x}            \\ \hline\hline
\end{tabular}%
}
\caption{Comparison of Si-SAN and SBT methods. Both methods only leverage AST information.}
\label{t6}
\end{table}

\subsection{Si-SAN vs. SAN}

To valid the effectiveness of Si-SAN, we gradually replace SAN layers in original Transformer with Si-SAN. Take Transformer model with Si-SAN proportion of 50\% as an instance, we replace the second, fourth and last three encoder layers with Si-SAN and do not apply structure-induced module.

The results of variant models with incremental proportions of Si-SAN layers are shown in Table \ref{t3}. Intuitively, all of the Transformers obtain improvements when equipped with Si-SAN layers. We can also see that SiT outperforms Transformer with similar proportion of Si-SAN, which proves the effectiveness of structure-induced module. However, it is surprising that Transformer with all 6 layers of Si-SAN still outperforms original Transformer even if it may be over-pruned. 

\subsection{Si-SAN vs. Sparse SAN}

To further valid our structure-based approach, we compare the performance of structure-induced attention with other sparse attention patterns, window attention in Longformer, ETC \citep{DBLP:journals/corr/abs-2004-05150, DBLP:journals/corr/abs-2004-08483} and random attention in BigBird \citep{zaheer2020big}. We depict different attention patterns in Figure \ref{4}. The default sequence length in SiT is 400, and then we set both $ w $ and $ r $ to 64 in window and random attention respectively.

As shown in Table \ref{t4}, Transformer with arbitrary sparse attention can not bring improvement as Si-SAN, which refutes that SiT learns better through denoising. Specifically, random attention seriously deteriorates Transformer. It is surprising that window attention achieves a better result than Vanilla Transformer. Intuitively, tree structures like AST are highly localized. That is why window attention may show good performance. Nevertheless, Transformer with Si-SAN still outperforms window attention by 0.52 BLEU point.

\begin{table}[]
\centering
\resizebox{0.49\textwidth}{!}{%
\begin{tabular}{@{}l|cccc@{}}
\hline\hline
\textbf{Model}                                                     & \textbf{Para.} & \textbf{BLEU} & \textbf{RE.L} & \textbf{MTR.} \\ \hline
Transformer-8    & 140M            & 47.42         & 57.28            & 29.62           \\
SiT-8            & 139M            & 50.04         & 59.56            & 31.46           \\
Transformer-12   & 244M            & 50.11         & 59.47            & 31.44           \\
SiT-12           & 242M            & 50.53         & 60.08            & 31.96           \\
Transformer-16   & 370M            & 50.43         & 59.80            & 31.75           \\
SiT-16           & 367M            & \textbf{50.97}         & \textbf{60.51}            & \textbf{32.35}  \\ \hline\hline
Transformer-ALBERT enc. & 124M           & 44.83         & 55.34            & 27.73           \\
SiT-ALBERT enc.         & 124M            & \textbf{49.31}         & \textbf{58.46}            & \textbf{30.83}           \\ \hline\hline
\end{tabular}%
}
\caption{BLEU, ROUGE-L and METEOR for variant models with different sizes, where RE.L and MTR. refer to ROUGE-L and METEOR respectively. Models like SiT-12 refers to SiT with 12 heads.}
\label{t5}
\end{table}

\subsection{Si-SAN vs. SBT}

We reproduce SBT method on Java \citep{hu2018deep} and apply it on our Transformer. For fair enough comparison, we let $\beta=\gamma=0$ and conduct single-view SiT which only leverages AST information. As depicted in Figure \ref{t6}, flattening ASTs into linear sequences does not result in improvement, which is consistent with \citet{ahmad-etal-2020-transformer}. However, we achieve substantial improvement while incorporating AST into Transformer using Si-SAN, which indicates our improved model design is indeed effective.

In addition, the average length of the input code will be much longer with SBT, which may introduce additional training cost. As shown in Figure \ref{t6}, SiT is 1.5 times faster than Transformer with SBT.

\subsection{Large Model}

It is known that for nearly all deep models, increasing model size may cover quite much of model structure design improvement. Thus, it is possible that the improvement on base-size model may not work on large-size one. To valid this, we compare SiTs with Transformers under larger scale. As we can see pictorially in Table \ref{t5}, with increasing parameter scale, SiTs with 12 heads and 16 heads both outperform the corresponding Transformers by 0.42 and 0.54 BLEU point respectively.

\subsection{Parameter Sharing}

Recently, parameter sharing on BERT \citep{DBLP:conf/naacl/DevlinCLT19} has achieved promising results \citep{DBLP:conf/iclr/LanCGGSS20}. Similar as ALBERT, we introduce cross-layer parameter sharing in both Transformer and SiT, sharing all parameters in all encoder layers. Note that we train our models from scratch and keep the decoder fixed.

As shown in Table \ref{t5}, SiT performs much better on parameter sharing than Transformer does. We believe that code summarization task highly depends on structural information, and this is why SiT can still achieve good results with simply one group of encoder parameters while Transformer encounters a serious decline. On the other hand, it makes possible for lite model, which may balance high efficiency and performance.

\section{Related Work}

\paragraph{RNN-based Approaches} While numbers of works \citep{haiduc2010use, eddy2013evaluating, wong2013autocomment, wong2015clocom, 10.1145/3377811.3380383} on code summarization usually depended on information retrieval, most of the recent works tend to treat it as a machine translation problem. Meanwhile attention mechanism is broadly used for better performance on capturing long-range features. \citet{pmlr-v48-allamanis16} proposed a Convolution Neural Network (CNN) with copy attention, and more commonly, \citet{DBLP:conf/acl/IyerKCZ16, DBLP:conf/aaai/LiangZ18} proposed to use Recurrent Neural Network (RNN) with attention mechanism to summarize code snippets into natural language. \citet{ijcai2018-314} introduced API knowledge from related tasks while \citet{DBLP:conf/acl/CaiLXLHC20} introduced type information to assist training, which also gained promising results. Additionally, reinforce learning \citep{wan2018improving} and dual learning \citep{wei2019code, DBLP:conf/www/YeXZHWZ20} are also shown effective to boost model performance.

\paragraph{Transformer-based Approaches} It is known that RNN-based models may encounter bottleneck when modeling long code sequences. \citet{ahmad-etal-2020-transformer} proposed an enhanced Transformer with copy attention and relative position encoding while \citet{DBLP:journals/corr/abs-2004-00998, DBLP:journals/corr/abs-2007-15813} proposed to use Transformer \citep{vaswani2017attention} and Transformer-XL \citep{DBLP:conf/acl/DaiYYCLS19}, all of which outperformed previous RNN-based models by a large margin.

\paragraph{Structure-based Approaches}
Recent works on code summarization pay more and more attention on structural information, which usually treats the source code in form of its Abstract Syntax Tree (AST). \citet{hu2018deep, DBLP:conf/icse/LeClairJM19,alon2018codeseq} leveraged flattened ASTs as inputs and trained with LSTMs. \citet{DBLP:conf/aaai/MouLZWJ16, bui2021treecaps, shido2019automatic, harer2019tree} proposed TBCNN, TreeCaps, Tree-LSTM and Tree-Transformer to directly encode tree-style inputs. Differ from modeling code with sequential models, \citet{allamanis2018learning, liu2020automatic, leclair2020codegnn} treated AST as graph and applied graph neural network, while \citet{wang2021learning} applied heterogeneous graph neural network to model different types of nodes.

\paragraph{Pre-training Approaches}Apart from training from scratch, CodeBERT \citep{feng2020codebert} is pre-trained on vast bimodal corpora with masked language model \citep{DBLP:conf/naacl/DevlinCLT19} and replaced token detection \citep{DBLP:conf/iclr/ClarkLLM20}, and achieves powerful performances on downstream tasks. \citet{DBLP:journals/corr/abs-2007-06934} intensified contextualized code representation through masked code fragment predictions while \citet{bui2020infercode} incorporated structural information using TBCNN. However, all of them do not include generation-related objectives. It is worth further exploration and practice on pre-training approaches for out concerned tasks.

\section{Conclusion}
This paper presents a novel structured-induced Transformer model on code summarization task. By well-designed architecture, the proposed model may effectively incorporate multi-view structure into attention mechanism without tricky implementation. We further adopt a new module architecture to aggregate both global self-attention and structure-induced self-attention representations. Experiments on two challenging benchmarks including Java and Python show that the proposed model yields new state-of-the-art results.

\bibliographystyle{acl_natbib}
\bibliography{acl2021}

\newpage
\appendix

\section{Model Parameters}
\label{b}

\begin{table}[h]
\centering
\resizebox{0.45\textwidth}{!}{%
\begin{tabular}{@{}l|ccccc@{}}
\hline\hline
Model              & $d_h$ & $d_{ff}$ & $h$ & $l$ \\ \hline
Transformer        & 64 & 2048 & 8  & 12      \\
SiT                & 64 & 2048 & 8  & 12      \\
Transformer-Window & 64 & 2048 & 8  & 12      \\
Transformer-Random & 64 & 2048 & 8  & 12      \\
Transformer-Struc. & 64 & 2048 & 8  & 12      \\
CodeBERT           & 64 & 3072 & 12 & 12      \\
SiT on CodeBERT    & 64 & 3072 & 12 & 12      \\
Transformer-ALBERT enc. & 64 & 2048 & 8  & 12      \\
SiT-ALBERT enc.    & 64 & 2048 & 8  & 12      \\ \hline\hline
\end{tabular}%
}
\caption{Model parameters in our experiments.}
\label{model}
\end{table}

\section{Qualitative Samples}
\label{c}

For qualitative analysis, we give some samples of code summarization with different models. We can see that SiT performs most precisely, while CodeBERT performs better than Transformer does.

\begin{table*}[]
\centering
\resizebox{0.8\textwidth}{!}{%
\begin{tabular}{l|l}
\hline\hline
Code    & \begin{tabular}[c]{@{}l@{}}\textcolor[rgb]{0.5,0.25,0.6}{private} \textcolor[rgb]{0.15,0.5,0.4}{float} \textcolor{blue}{computeOverscrollPercent} () \{\\
\quad\; \textcolor[rgb]{0.5,0.25,0.6}{if} ( mOverScrollOffset \textgreater{}= \_NUM ) \{\textcolor[rgb]{0.5,0.25,0.6}{return} mOverScrollOffset / mMaxOverScroll;\}\\
\quad\; \textcolor[rgb]{0.5,0.25,0.6}{else} \{\textcolor[rgb]{0.5,0.25,0.6}{return} mOverScrollOffset / mMaxUnderScroll;\}\\
\}\end{tabular}\\ \hline
Summary & \begin{tabular}[c]{@{}l@{}}\textcolor[rgb]{0.7,0.6,0.1}{Gold}: determine the current amount of overscroll . if the value is 0 , there is no overscroll . if the value is $ < $ 0 , tabs are\\ overscrolling towards the top or or left . if the value is $ > $ 0 , tabs are overscrolling towards the \textcolor[rgb]{0.8,0.2,0.2}{bottom} or right .\\
\textcolor[rgb]{0.8,0.2,0.2}{SiT}: determine the current amount of overscroll . if the value is 0 , there is no overscroll . if the value is $ < $ 0 , tabs are\\ overscrolling towards the \textcolor[rgb]{0.8,0.2,0.2}{top} or or left . if the value is $ > $ 0 , tabs are overscrolling towards the \textcolor[rgb]{0.8,0.2,0.2}{bottom} or right .\\
\textcolor{blue}{Transformer}: determine the current amount of overscroll . if the value is 0 , there is no overscroll . if the value is $ < $ 0 , tabs\\ are overscrolling towards the top or or left . if the value is $ > $ 0 , tabs are overscrolling towards the top or right .\\
\textcolor[rgb]{0.5,0.25,0.6}{CodeBERT}: determine the current amount of overscroll . if the value is 0 , there is no overscroll . if the value is $ < $ 0 , tabs\\ are overscrolling towards the top or or left . if the value is $ > $ 0 , tabs are overscrolling towards the \textcolor[rgb]{0.8,0.2,0.2}{bottom} or right .\end{tabular}\\ \hline\hline
Code    & \begin{tabular}[c]{@{}l@{}}\textcolor[rgb]{0.5,0.25,0.6}{public} \textcolor[rgb]{0.15,0.5,0.4}{String} \textcolor{blue}{peek} () \{\\
\quad\; \textcolor[rgb]{0.15,0.5,0.4}{String} result = null;\\
\quad\; \textcolor[rgb]{0.5,0.25,0.6}{if} (isEmpty()) \{\textcolor[rgb]{0.5,0.25,0.6}{return} null;\}\\
\quad\; \textcolor[rgb]{0.5,0.25,0.6}{else} \{\\
\quad\; \quad\; \textcolor[rgb]{0.15,0.5,0.4}{int} cachedCurrentIndex = currentIndex;\\
\quad\; \quad\; \textcolor[rgb]{0.5,0.25,0.6}{if} (isEatingBlocksOfDelimiters) \{trimStartingDelimiters();\}\\
\quad\; \quad\; \textcolor[rgb]{0.15,0.5,0.4}{int} nearestDelimeter = -\_NUM ;\\
\quad\; \quad\; \textcolor[rgb]{0.5,0.25,0.6}{for} (\textcolor[rgb]{0.15,0.5,0.4}{int} i = \_NUM; i \textless delimiters.length(); i++) \{\\
\quad\; \quad\; \quad\; \textcolor[rgb]{0.15,0.5,0.4}{int} delimiter = source.indexOf(delimiters.charAt(i), currentIndex);\\
\quad\; \quad\; \quad\; \textcolor[rgb]{0.5,0.25,0.6}{if} (nearestDelimeter == -\_NUM $ \mid \mid $ delimiter != -\_NUM \&\& delimiter \textless nearestDelimeter) \{\\
\quad\; \quad\; \quad\; \quad\; nearestDelimeter = delimiter;\\
\quad\; \quad\; \quad\; \}\\
\quad\; \quad\; \}\\
\quad\; \quad\; \textcolor[rgb]{0.5,0.25,0.6}{if} (nearestDelimeter == -\_NUM) \{result = source.substring(currentIndex);\}\\
\quad\; \quad\; \textcolor[rgb]{0.5,0.25,0.6}{else} \{result = source.substring(currentIndex, nearestDelimeter);\}\\
\quad\; \quad\; currentIndex = cachedCurrentIndex;\\
\quad\; \}\\
\quad\; \textcolor[rgb]{0.5,0.25,0.6}{return} result;\\
\}\end{tabular} \\ \hline
Summary & \begin{tabular}[c]{@{}l@{}}\textcolor[rgb]{0.7,0.6,0.1}{Gold}: returns null if there is nothing left .\\
\textcolor[rgb]{0.8,0.2,0.2}{SiT}: returns null if there is nothing left .\\
\textcolor{blue}{Transformer}: finds the next unique identifier .\\
\textcolor[rgb]{0.5,0.25,0.6}{CodeBERT}: returns the index of the first delimited string removing from the current position .\end{tabular}\\ \hline\hline
\end{tabular}%
}
\caption{Qualitative samples of Java code summarization.}
\label{java}
\end{table*}

\begin{table*}[]
\centering
\resizebox{0.8\textwidth}{!}{%
\begin{tabular}{l|l}
\hline\hline
Code    & \begin{tabular}[c]{@{}l@{}}\textcolor[rgb]{0.5,0.25,0.6}{def}
\textcolor{blue}{\_asFilesystemBytes}(path, encoding=\textcolor[rgb]{0.5,0.25,0.6}{None}):\\
\quad\; \textcolor[rgb]{0.5,0.25,0.6}{if} (type(path) == bytes): \textcolor[rgb]{0.5,0.25,0.6}{return} path\\
\quad\; \textcolor[rgb]{0.5,0.25,0.6}{else}:\\
\quad\; \quad\; \textcolor[rgb]{0.5,0.25,0.6}{if} (encoding \textcolor[rgb]{0.5,0.25,0.6}{is None}):\\
\quad\; \quad\; \quad\; encoding = sys.getfilesystemencoding()\\
\quad\; \quad\; \quad\; \textcolor[rgb]{0.5,0.25,0.6}{return} path.encode(encoding)\end{tabular}  \\ \hline
Summ. & \begin{tabular}[c]{@{}l@{}}\textcolor[rgb]{0.7,0.6,0.1}{Gold}: return c{path} as a string of l{bytes} suitable for use on this systems filesystem .\\
\textcolor[rgb]{0.8,0.2,0.2}{SiT}: return c{path} as a string of l{unicode} suitable for use on this systems filesystem .\\
\textcolor{blue}{Transformer}: convert a filesystem path of a byte string .\\
\textcolor[rgb]{0.5,0.25,0.6}{CodeBERT}: return a byte string suitable for use in c{path} as a byte string .\end{tabular}\\ \hline\hline
Code    & \begin{tabular}[c]{@{}l@{}}\textcolor[rgb]{0.5,0.25,0.6}{def} \textcolor{blue}{absent}(name, DomainName, region=\textcolor[rgb]{0.5,0.25,0.6}{None}, key=\textcolor[rgb]{0.5,0.25,0.6}{None}, keyid=\textcolor[rgb]{0.5,0.25,0.6}{None}, profile=\textcolor[rgb]{0.5,0.25,0.6}{None}):\\
\quad\; ret = \{\textcolor[rgb]{0.8,0.25,0.25}{'name'}: DomainName, \textcolor[rgb]{0.8,0.25,0.25}{'result'}: \textcolor[rgb]{0.5,0.25,0.6}{True}, \textcolor[rgb]{0.8,0.25,0.25}{'comment'}: '', \textcolor[rgb]{0.8,0.25,0.25}{'changes'}: \{\}\}\\
\quad\; r = \_\_salt\_\_{[}'boto\_elasticsearch\_domain.exists'{]}(DomainName, region=region, key=key, keyid=keyid, profile=profile)\\
\quad\; \textcolor[rgb]{0.5,0.25,0.6}{if} (\textcolor[rgb]{0.8,0.25,0.25}{'error'} \textcolor[rgb]{0.5,0.25,0.6}{in} r):\\
\quad\; \quad\; ret{[}\textcolor[rgb]{0.8,0.25,0.25}{'result'}{]} = \textcolor[rgb]{0.5,0.25,0.6}{False}\\
\quad\; \quad\; ret{[}\textcolor[rgb]{0.8,0.25,0.25}{'comment'}{]} = \textcolor[rgb]{0.8,0.25,0.25}{'Failed to delete domain: \{0\}.'}.format(r{[}\textcolor[rgb]{0.8,0.25,0.25}{'error'}{]}{[}\textcolor[rgb]{0.8,0.25,0.25}{'message'}{]})\\
\quad\; \quad\; \textcolor[rgb]{0.5,0.25,0.6}{return} ret\\
\quad\; \textcolor[rgb]{0.5,0.25,0.6}{if} (r \textcolor[rgb]{0.5,0.25,0.6}{and} (\textcolor[rgb]{0.5,0.25,0.6}{not} r{[}'exists'{]})):\\
\quad\; \quad\; ret{[}\textcolor[rgb]{0.8,0.25,0.25}{'comment'}{]} = \textcolor[rgb]{0.8,0.25,0.25}{'Domain \{0\} does not exist.'}.format(DomainName)\\
\quad\; \quad\; \textcolor[rgb]{0.5,0.25,0.6}{return} ret\\
\quad\; \textcolor[rgb]{0.5,0.25,0.6}{if} \_\_opts\_\_{[}'test'{]}:\\
\quad\; \quad\; ret{[}\textcolor[rgb]{0.8,0.25,0.25}{'comment'}{]} = \textcolor[rgb]{0.8,0.25,0.25}{'Domain \{0\} is set to be removed.'}.format(DomainName)\\
\quad\; \quad\; ret{[}\textcolor[rgb]{0.8,0.25,0.25}{'result'}{]} = \textcolor[rgb]{0.5,0.25,0.6}{None} \\
\quad\; \quad\; \textcolor[rgb]{0.5,0.25,0.6}{return} ret\\
\quad\; r = \_\_salt\_\_{[}\textcolor[rgb]{0.8,0.25,0.25}{'boto\_elasticsearch\_domain.delete'}{]}(DomainName, region=region, key=key, keyid=keyid, profile=profile)\\
\quad\; \textcolor[rgb]{0.5,0.25,0.6}{if} (\textcolor[rgb]{0.5,0.25,0.6}{not} r{[}'deleted'{]}):\\
\quad\; \quad\; ret{[}\textcolor[rgb]{0.8,0.25,0.25}{'result'}{]} = \textcolor[rgb]{0.5,0.25,0.6}{False}\\
\quad\; \quad\; ret{[}\textcolor[rgb]{0.8,0.25,0.25}{'comment'}{]} = \textcolor[rgb]{0.8,0.25,0.25}{'Failed to delete domain: \{0\}.'}.format(r{[}\textcolor[rgb]{0.8,0.25,0.25}{'error'}{]}{[}\textcolor[rgb]{0.8,0.25,0.25}{'message'}{]})\\
\quad\; \quad\; \textcolor[rgb]{0.5,0.25,0.6}{return} ret\\
\quad\; ret{[}\textcolor[rgb]{0.8,0.25,0.25}{'changes'}{]}{[}\textcolor[rgb]{0.8,0.25,0.25}{'old'}{]} = \{\textcolor[rgb]{0.8,0.25,0.25}{'domain'}: DomainName\}\\
\quad\; ret{[}\textcolor[rgb]{0.8,0.25,0.25}{'changes'}{]}{[}\textcolor[rgb]{0.8,0.25,0.25}{'new'}{]} = \{\textcolor[rgb]{0.8,0.25,0.25}{'domain'}: \textcolor[rgb]{0.5,0.25,0.6}{None}\}\\
\quad\; ret{[}\textcolor[rgb]{0.8,0.25,0.25}{'comment'}{]} = \textcolor[rgb]{0.8,0.25,0.25}{'Domain \{0\} deleted.'}.format(DomainName)\\
\quad\; \textcolor[rgb]{0.5,0.25,0.6}{return} ret\end{tabular} \\ \hline
Summ. & \begin{tabular}[c]{@{}l@{}}\textcolor[rgb]{0.7,0.6,0.1}{\textcolor[rgb]{0.7,0.6,0.1}{Gold}}: ensure domain with passed properties is absent .\\
\textcolor[rgb]{0.8,0.2,0.2}{SiT}: ensure domain with passed properties is absent .\\
\textcolor{blue}{Transformer}: ensure the iam role exists .\\
\textcolor[rgb]{0.5,0.25,0.6}{CodeBERT}: ensure the named domain is absent .\end{tabular}\\ \hline\hline
\end{tabular}%
}
\caption{Qualitative samples of Python code summarization.}
\label{java}
\end{table*}

\end{document}